\pdfoutput=1  
\documentclass[conference]{IEEEtran}
\IEEEoverridecommandlockouts

\usepackage{cite}
\usepackage{graphicx}
\usepackage{amsmath,amssymb,amsfonts}
\usepackage{algorithm}
\usepackage{algpseudocode}
\usepackage{xcolor}
\usepackage{booktabs}
\usepackage{url}
\usepackage{textcomp}

\newcommand{\blfootnote}[1]{%
  \begingroup
  \renewcommand\thefootnote{}\footnote{#1}%
  \addtocounter{footnote}{-1}%
  \endgroup
}

\begin{document}

\title{AutoResearch at Production Scale: Failure Modes and a Multi-Agent Framework}

\author{%
\centering
\begin{tabular}{c@{\hspace{3em}}c@{\hspace{3em}}c}
\shortstack{Aparajith Chandran\\[2pt]\textit{Books Core Recs Science}\\\textit{Amazon}\\Seattle, WA, USA\\aparajic@amazon.com} &
\shortstack{Juwon Kim\\[2pt]\textit{Books Core Recs Science}\\\textit{Amazon}\\Seattle, WA, USA\\juwonki@amazon.com} &
\shortstack{Saurav Jha\\[2pt]\textit{Books Core Recs Science}\\\textit{Amazon}\\Seattle, WA, USA\\sauravj@amazon.com} \\[1.4em]
\multicolumn{3}{c}{%
  \shortstack{Pablo Castells\\[2pt]\textit{Books Core Recs Science}\\\textit{Amazon}\\Seattle, WA, USA\\pcste@amazon.com}%
  \hspace{3em}%
  \shortstack{Florian Hottier\\[2pt]\textit{Books Core Recs Science}\\\textit{Amazon}\\Seattle, WA, USA\\hottier@amazon.com}%
} \\
\end{tabular}%
}

\maketitle

\blfootnote{\copyright~2026 IEEE. Personal use of this material is permitted. Permission from IEEE must be obtained for all other uses, in any current or future media, including reprinting/republishing this material for advertising or promotional purposes, creating new collective works, for resale or redistribution to servers or lists, or reuse of any copyrighted component of this work in other works. Accepted for publication in the Proceedings of the 2026 IEEE International Conference on Data Mining (ICDM).}

\begin{abstract}
Optimizing embedding systems for production recommendation pipelines demands systematic exploration that consumes disproportionate engineering effort at scale. We apply Andrej Karpathy's AutoResearch paradigm~\cite{karpathy2026autoresearch}---a large language model that iteratively edits a training script and retains modifications that improve a held-out scalar metric---to automate this exploration. We report on twelve weeks of running this paradigm at production scale, where iterations consume hours of multi-GPU compute, evaluation involves competing criteria, and campaigns span weeks across many training jobs. Across two independently developed representation-learning systems for a book recommendation pipeline, we ran 220+ experiments and observed five recurring failure modes absent from the original setting: infrastructure fragility, agent memory decay, search-direction stagnation, iteration-cost asymmetry, and metric fixation. We contribute a three-principle scaffolding design---prevent, persist, redirect---that maps each failure mode to a structural remedy and whose instantiation scales with iteration cost. The framework produced a 1.82$\times$ Recall@6 lift and a 2.1$\times$ coherence lift over hand-tuned baselines, and the agent autonomously designed a text-only fallback that expanded catalog coverage by 5.8$\times$. The two systems span nearly three orders of magnitude in per-iteration cost yet exhibit the same failure modes, suggesting these are structural properties of production-scale autonomous research rather than artifacts of either application.
\end{abstract}


\section{Introduction}

Andrej Karpathy's AutoResearch~\cite{karpathy2026autoresearch}---a 630-line Python script---introduced an elegant paradigm for autonomous machine learning research: a large language model iteratively modifies a training script, runs the experiment, and keeps or discards based on a held-out scalar metric. Unlike AutoML approaches~\cite{akiba2019optuna} that pick from a predefined search space, AutoResearch proposes and modifies arbitrary code: architecture, optimizer, loss function, training tricks. Applied to single-GPU, five-minute language-modeling experiments, it demonstrated that code-level iterative search can discover meaningful improvements without human intervention.

The paradigm has obvious appeal for production ML, where exploration budgets are large but engineering attention is scarce~\cite{sculley2015debt}. Yet production ML differs from the original setting along three axes that strain the paradigm's assumptions. First, production scale often results in iterations consuming hours of multi-GPU compute rather than minutes, raising the cost of any failure by three orders of magnitude. Second, evaluation involves complex, competing criteria rather than a single scalar, so ``keep if metric improved'' is no longer sufficiently defined. Third, campaigns span weeks across many training jobs, exceeding any single context window. Whether AutoResearch survives these conditions has, to our knowledge, not been reported.

This paper reports on twelve weeks of running AutoResearch at production compute scale within Amazon's book recommendation pipeline. We deployed the paradigm in two independent systems with very different cost regimes. \textbf{System~A} learns 128-dimensional purchase optimized retrieval embeddings for 98\% of the active catalog on multiple GPU instances (consuming 9--19 hour iterations); the agent rewrites the entire training script per iteration. \textbf{System~B} learns hierarchical Semantic IDs for books via Residual Quantization (RQ-VAE)~\cite{zeghidour2021soundstream,vandenoord2017vqvae} used for generative retrieval~\cite{rajput2024genret} and taxonomy on a single GPU (6--52 minute iterations); the agent edits hyperparameters and model architecture rather than the full script. The systems span nearly three orders of magnitude in per-iteration cost and were developed independently by different workstreams, which makes their shared findings more compelling. Both systems produced significant improvements over hand-tuned baselines: 1.82$\times$ Recall@6 lift on System~A, and 2.1$\times$ lift on the custom evaluation metric used for System~B. In terms of performance, there was a 5.8$\times$ catalog-coverage scale-up on System~A, designed and implemented by the agent itself with no human direction.

\textbf{Related work.} Several recent systems explore autonomous experimentation and iterative LLM-driven code search. The AI Scientist~\cite{lu2024aiscientist} targets full-paper authorship via batch-style hypothesis generation and writeup, working at a coarser granularity than AutoResearch. ScientistOne~\cite{meng2026scientistone} extends this line toward end-to-end autonomous research with verifiable evidence chains, but is evaluated on benchmark research tasks where iteration is cheap and the output is a paper rather than a production system. MLAgentBench~\cite{huang2024mlagentbench} evaluates language agents on standardized small-scale ML tasks, focusing on benchmarking rather than deployment. Self-Refine~\cite{madaan2023selfrefine} and Reflexion~\cite{shinn2023reflexion} formalize self-correction loops where an LLM critiques and revises its own output, but operate within a single context window. LLMs as Optimizers~\cite{yang2024optimizers} and EvoPrompting~\cite{chen2023evoprompting} treat the LLM as a black-box optimizer over prompts or architectures. FunSearch~\cite{romeraparedes2024funsearch} couples a language model with an evolutionary loop for mathematical discovery. AIDE~\cite{jiang2025aide} frames ML engineering as tree-search over code and achieves strong benchmark performance; SWE-agent~\cite{yang2024sweagent} builds specialized agent-computer interfaces for repository-level code tasks. Both report on structured benchmark settings rather than production deployment. Multi-agent coordination frameworks such as ReAct~\cite{yao2023react} and AutoGen~\cite{wu2023autogen} address agent communication and tool use. MemGPT~\cite{packer2023memgpt} addresses context-length limits via hierarchical memory management. None of these systems reports what happens when iteration cost is high, evaluation is multi-criteria, and a campaign exceeds a single context. Our work occupies that gap.

\textbf{Contributions.} This paper makes three contributions: (1) \textbf{a taxonomy of five failure modes of AutoResearch} that emerge at production scale---infrastructure fragility, agent memory decay, search-direction stagnation, iteration-cost asymmetry, and metric fixation---each with an identified root cause, with the first four independently observed across two systems of very different cost profiles and the fifth problematic only to multi-criteria evaluation; (2) \textbf{a three-principle scaffolding design} --- prevent\textit{ }(pre-execution semantic gate),\textit{ }persist\textit{ }(durable cross-job memory)\textit{, }redirect (stagnation-triggered strategic reframing) --- that maps each failure mode to a structural remedy, instantiated as a three-agent extension of AutoResearch and validated against a 1.82× Recall@6 lift and a 2.1× coherence lift across the two systems; and (3) a cost-dependence principle: \textbf{the weight placed on each scaffolding principle should scale with iteration cost} - pre-execution gates justify their overhead only when iterations are expensive; post-execution revert suffices when they are cheap. This principle, observed empirically across two systems differing from each other by nearly three orders of magnitude in cost, offers a concrete design parameter for teams deploying autonomous research agents beyond the settings studied here. We note that the individual mechanisms we employ---persistent memory, critique loops, pre-execution gating---are deliberately standard; the contribution is not a new algorithm but a failure taxonomy grounded in production deployment and a cost-dependence principle governing when each mechanism earns its overhead.

\section{Production Setting}

Both systems share the standard AutoResearch components~\cite{karpathy2026autoresearch}: 1) an orchestrator, a research program (\texttt{program.md}) carrying instructions and accumulated findings; 2) an agent sandbox containing the training script (\texttt{train.py}), an immutable evaluator that the agent observes but cannot modify; and 3) a research log of all iterations. They differ in adaptation, scale, and cost regime (Table~\ref{tab:comparison}). We introduce the two systems in the following paragraphs.

\subsection{System A: Embedding Generation}

System~A generates retrieval embeddings for 98\% of the active catalog, translating to tens of millions of books. For the purpose of this study, we use \textbf{Recall@6} as the objective metric: for each held-out book a customer purchased, the six nearest neighbors in embedding space are retrieved via a FAISS~\cite{johnson2019faiss} IVF+Flat index, and Recall@6 measures how often these similarity-based recommendations include the customer's actual next purchase. Top 6 is a representative approximation of visible items in the recommender's interface. The input data comprises a co-purchase behavioral graph suitable for Node2Vec-style~\cite{grover2016node2vec} training and pre-computed text embeddings from a sentence transformer~\cite{reimers2019sbert}. The hand-tuned baseline against which the agent is compared is the product of four engineer-weeks of manual work: a Node2Vec implementation on the co-purchase graph, distributed training on GPUs, and iteration over loss functions and negative sampling, reaching 3.79\% Recall@6.

\subsection{System B: Configuration-Level Semantic ID Assignment}

System~B learns hierarchical Semantic IDs~\cite{rajput2024genret} for over 50\% of browsed books (in the order of millions) via RQ-VAE~\cite{vandenoord2017vqvae,zeghidour2021soundstream}, which encodes each book's continuous embedding into a sequence of discrete codes, each level quantizing the residual left by
the previous one. Books that agree on their first code fall in the same coarse cluster; agreeing on the first two codes places them in a finer sub-cluster, and so on---each additional code is a deeper level of a browse-path-like hierarchy. Our primary metric, \textbf{weighted coherence}, measures how similar books within a cluster are. Writing $L1$, $L2$, $L3$ for the mean cosine similarity between book embeddings within clusters at the first three levels, weighted coherence is $0.5 \times L1 + 0.3 \times L2 + 0.2 \times L3$. The weights reflect customer-flow dominance: top-level prefix matches are the most frequent shared-prefix event across the catalog and most directly capture browsable cluster structure, so $L1$ carries half the weight, while $L3$ contributes the smallest share because deeper levels fragment quickly with cohort size. The weights are fixed across all System~B runs and were not tuned to any individual configuration. 

We also use secondary metrics approximating genre purity and cluster usability (median size, singleton rate, collision rate, etc.). The agent operates at the configuration and architecture level---proposing changes to hyperparameters and encoder architecture but not rewriting the training script. The hand-tuned baseline is a manually configured RQ-VAE trained on behavioral embeddings with hand-tuned codebook size, learning rate, and commitment weight (2--3 days of prototyping), reaching weighted coherence of 0.348---already well above the random-assignment baseline weighted coherence of 0.26.

\subsection{Cost Profile and Campaign Scale}

The combined campaign spans twelve weeks and 220+ experiments. System~A ran 60+ iterations across 6 runs; System~B ran 150+ iterations across 15 runs. A single System~A iteration costs roughly 760$\times$ a System~B iteration (combining instance hourly price, GPU count, and wall-clock duration); the two thus differ in per-iteration cost by nearly three orders of magnitude, making them an unusually informative pair for studying cost-dependent design choices.

\begin{table}[tbp]
  \caption{System comparison. The two systems span nearly three orders of magnitude in per-iteration cost ($\sim$760$\times$).}
  \label{tab:comparison}
  \centering
  \resizebox{\columnwidth}{!}{%
  \begin{tabular}{lll}
    \toprule
    Dimension & System A & System B \\
    \midrule
    Search level & Code-level (\texttt{train.py}) & Config + arch.\ \\
    Iteration wall-clock & 9--19 hours  & 6--52 min, 1 GPU \\
    Cost per iteration & $\sim$760$\times$ System B & baseline \\
    \# iterations & 60+ across 6 runs & 150+ across 15 runs \\
    Metric & Single scalar (R@6) & Multi-criteria \\
    Steering & Criticizer (automated) & \texttt{program.md} (human) \\
    Agents & 3 (Researcher, Fixer, Critic) & 1 (Researcher) \\
    \bottomrule
  \end{tabular}}
\end{table}

\section{Failure Modes at Production Scale}

We organize the failure modes by their root cause, distinguishing modes that follow from \textit{cost} (A, D) from those that follow from \textit{long horizons} (B, C) and those that follow from \textit{evaluation design} (E). The first four modes were observed independently in both systems despite their different cost profiles; metric fixation (mode 5) requires multi-criteria evaluation and is therefore specific to System~B, whose primary metric competes with secondary usability criteria (System~A optimizes a single scalar and cannot exhibit it).

\subsection{Infrastructure Fragility}

Iterations may fail from environmental reasons---evaluation timeouts, out-of-memory (OOM) crashes, GPU distribution bugs, malformed inter-agent messages---consuming budget without producing a usable signal. In System~A, a GraphSAGE model exceeded the memory of multiple GPUs, and---before the Code Fixer was introduced---a multi-GPU wrapper bug ran 70M text encodings on 1 of the GPUs for over 10 hours. This class of fault is exactly what later motivated the pre-execution Code Fixer; once deployed, it caught recurrences of the same DDP misconfiguration before they reached hardware (Section~IV-B). In System~B, 11 of 30 iterations in a single run failed because the evaluator could not process a 10$\times$ batch size of semantic IDs within its 600-second timeout; eleven consecutive iterations were wasted before a human intervened. Coordination fragility is invisible in single-agent settings: it emerges only when multiple agents communicate through structured protocols and either end can silently corrupt the other's input. \textit{Root cause:} the agent has no model of infrastructure constraints, and agent-to-agent channels can degrade without either end recognizing the breakdown.

\subsection{Agent Memory Decay}

The agent re-tests configurations that have already failed, sometimes repeatedly, rationalizing each re-test as ``this time the context is different.'' In System~B, training for 100 epochs was tested and failed (due to overfitting) at least six times across multiple runs, despite \texttt{program.md} explicitly stating ``100 epochs causes overfitting---do not re-test.'' System~A showed the same pattern at code level: in one run the Researcher re-introduced the same prohibited file-system operation six times across four iterations before any training completed, and across the campaign roughly five iterations were lost re-exploring configurations already recorded as failures. \textit{Root cause:} a limited context window prevents holding full history; no persistent memory of negative results across runs; and the agent's tendency to rationalize, consistent with broader observations of LLM confabulation~\cite{ji2023hallucination}; external memory mechanisms such as MemGPT~\cite{packer2023memgpt} reduce but do not eliminate this pattern.

\subsection{Search-Direction Stagnation}

The agent reaches a local optimum and continues making marginal tweaks producing changes within noise, without recognizing the search direction is exhausted. In System~A, without external redirection, the Researcher would have continued fusing pre-compressed 128d features indefinitely. In System~B, one run spent five consecutive iterations attempting to improve a previous architecture, all within $\pm$0.008 weighted coherence. Another tested commitment weight at four values producing results within 0.01 of each other. \textit{Root cause:} the agent lacks a meta-level view of whether the current search direction has diminishing returns.

\subsection{Iteration-Cost Asymmetry}

The agent selects experiments without weighing expected information 
gain against compute cost, treating a 15-hour multi-GPU run as 
no more consequential than a 30-minute ablation. This is distinct 
from infrastructure fragility: fragility describes \textit{why} iterations fail; cost asymmetry describes \textit{how} the agent decides what to run in the first place. In System~A, the agent proposed a GraphSAGE variant requiring significantly more GPU memory than the baseline without flagging the compute overhead---the expected gain did not justify the marginal cost, but the agent had no mechanism to recognize this. In System~B, a behavior-based auxiliary loss was explored across 10 consecutive iterations with negligible genre purity improvement before the run was terminated; each iteration was cheap, but the accumulated cost of an unproductive search direction was not. A separate failure---catastrophic codebook collapse dropping weighted coherence from 0.562 to 0.503 from a single parameter change---illustrates a related asymmetry: the agent assigned equal prior probability to high-consequence and low-consequence changes, with no mechanism to hedge against irreversible moves. Root cause: the agent has no model of information gain relative to compute expenditure, and no prior over the consequence severity of proposed changes.

\subsection{Metric Fixation}

The agent optimizes the given metric faithfully but never questions whether it is the right metric, whether targets are feasible, whether the tradeoffs to meet the metric are reasonable, or whether the evaluation design has flaws. In System~B, the agent discovered that maximizing codebook size trivially improves weighted coherence by creating more fine-grained clusters, but the resulting hierarchy was unusable---deep levels collapsed into singletons. We had to introduce a guardrail constraining \texttt{n\_levels} and add a tertiary cluster-usability metric to prevent the agent from gaming coherence at the expense of a meaningful hierarchy. More strikingly, even with these additions, the three usability criteria (weighted coherence, genre purity, and median cluster size) were never simultaneously met in any of 15 runs across 150+ iterations, indicating that the tradeoff was structural rather than a tuning problem. While System A did not observe metric fixation as a failure due to it having a single optimization target, production systems will often require more complex, multi-objective targets.  \textit{Root cause:} the agent treats the metric as ground truth, not as a design choice.

\begin{table}[tbp]
  \caption{Failure-mode incidence across 220+ iterations. The four cost- and horizon-driven modes appear in both systems; metric fixation requires multi-criteria evaluation and is specific to System~B. Cells report the natural unit per mode (campaign-wide rate, incident count, or worst-run rate where noted); severity differs with iteration cost.}
  \label{tab:failure-frequency}
  \centering
  \begin{tabular}{lll}
    \toprule
    Failure mode & System A (60+ iter) & System B (150+ iter) \\
    \midrule
    Infra.\ fragility    & 18\% iter   & 37\% (worst run) \\
    Memory decay         & ${\sim}$5 inc.\ & ${\geq}6$ inc.\ \\
    Stagnation           & 5+ iter      & 5+ iter \\
    Cost asymmetry       & Cost wasted per crash & 10 iter wasted \\
    Metric fixation      & N/A          & 0/15 runs met \\
    \bottomrule
  \end{tabular}
\end{table}

The five modes share a common origin: small-scale experiments have cheap iterations, simple metrics, short sessions that fit in context, and negligible failure costs. At production scale these assumptions break, and the failure modes emerge. Table~\ref{tab:failure-frequency} aggregates incidence.

\section{Framework Design}

We organize our scaffolding around three principles: \textit{prevent} (stop failures before compute waste), \textit{redirect} (steer the agent when it stagnates), and \textit{persist} (maintain knowledge across iterations and runs). Each principle addresses a specific subset of the failure modes from Section~III: prevention targets infrastructure fragility and iteration-cost asymmetry; redirection targets search-direction stagnation; persistence targets memory decay. Metric fixation is only partially mitigated, as we discuss at the end of this section. Table~\ref{tab:scaffolding} summarizes how each principle is instantiated in the two systems. The principles are not specific to this application: any long-horizon agent loop with asymmetric failure costs, context resets between sessions, and a search space with local optima faces the same structural pressures. The weight placed on each principle should vary with iteration cost—a parameter we return to in Section~VI.

\begin{table}[t]
  \caption{The three scaffolding principles and their cost-appropriate
  instantiation in each system. The same principles apply across both
  systems; implementation differs with iteration cost.}
  \label{tab:scaffolding}
  \centering
  \resizebox{\columnwidth}{!}{%
  \begin{tabular}{lll}
    \toprule
    Principle & System A (high cost) & System B (low cost) \\
    \midrule
    Prevent  & Code Fixer agent (pre-execution) & Automatic revert (post-execution) \\
    Redirect & Criticizer agent (automated)     & Human cross-run pivot \\
    Persist  & S3 cross-job storage             & Human-maintained \texttt{program.md} \\
    \bottomrule
  \end{tabular}}
\end{table}

The implementation differs with iteration cost---pre-execution gates 
justify their overhead only when a failed iteration wastes hours of 
compute; post-execution revert suffices when iterations are cheap 
enough to treat as disposable hypotheses.

\subsection{Architecture}

\begin{figure*}[t]
  \centering
  \includegraphics[width=0.85\textwidth]{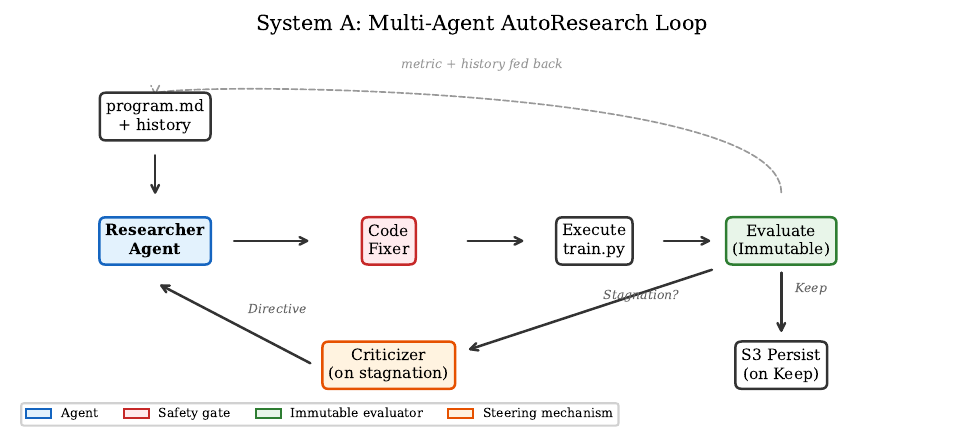}
  \caption{System~A three-agent architecture. The Researcher (Sonnet) produces a complete training script per iteration. The Code Fixer (Opus) rewrites the script pre-execution to repair latent infrastructure faults. The Criticizer (Opus) fires only on stagnation and emits a single strategic directive. The evaluator and \texttt{program.md} are immutable to the agents.}
  \label{fig:system-a-arch}
\end{figure*}

System~A extends AutoResearch with three agents: a \textbf{Researcher} (\texttt{claude-sonnet-4-6}), and a \textbf{Criticizer} and \textbf{Code Fixer} (\texttt{claude-opus-4-6}). The model split is intentional: Sonnet for the high-volume code-generation path; Opus for the lower-frequency, higher-leverage roles where a wrong call wastes hours of multi-GPU compute. Context management uses a sliding window with the most recent iteration summaries included verbatim and the current best \texttt{train.py} always available in full. Figure~\ref{fig:system-a-arch} shows the control flow; Algorithm~\ref{alg:loop} states it explicitly.

\begin{algorithm}[tbp]
\caption{Three-agent AutoResearch loop (System~A)}
\label{alg:loop}
\begin{algorithmic}[1]
\State Initialize \textit{best\_score}, $\textit{program} \gets$ Karpathy's \texttt{program.md}
\State $\textit{directive} \gets \varnothing$
\For{$i = 1, 2, \ldots$}
  \State $\textit{code} \gets$ \textsc{Researcher}(\textit{program}, \textit{history}, \textit{directive})
  \State $\textit{code} \gets$ \textsc{CodeFixer}(\textit{code}) \Comment{pre-execution rewrite}
  \State $\textit{score} \gets$ \textsc{Evaluator}(\textsc{Train}(\textit{code}))
  \If{$\textit{score} > \textit{best\_score} \cdot (1 + \tau_{\text{keep}})$}
    \State $\textit{best} \gets \textit{code}$; $\textit{best\_score} \gets \textit{score}$;
    \State $\textit{directive} \gets \varnothing$
  \EndIf
  \If{stagnant: $\Delta_{\text{rel}}(\textit{best\_score})_{[i-5, i]} < \tau_{\text{stag}}$}
    \State $\textit{directive} \gets$ \textsc{Criticizer}(\textit{history})
  \EndIf
  \State $\textit{program} \gets$ \textsc{Update}(\textit{program}, \textit{score}, \textit{summary})
\EndFor
\end{algorithmic}
\end{algorithm}

The \textbf{Researcher} receives a structured prompt containing the task specification, current best code, past iteration summaries, and optionally a strategic directive from the Criticizer. It produces a complete, self-contained training script with full freedom in architecture, loss function, and optimization strategy.

\subsection{Prevention: Code Fixer}

The Code Fixer reviews each Researcher-produced script before execution, catching semantic bugs that syntactic validation cannot detect: GPU underutilization, distributed-training misconfigurations, unnecessary recomputation, and memory leaks. Critically, the Code Fixer \textit{rewrites} the proposed code in a single pass rather than describing bugs in natural language for a separate authoring step. The single-pass design was a response to an earlier two-step Reviewer-based architecture that we abandoned: across one campaign, the Reviewer returned malformed JSON on 14 consecutive calls, and the Researcher re-introduced the same prohibited file-system operation six times across four iterations before any training completed. The Code Fixer eliminates this two-step authoring loop. In its first deployment, it corrected 14 latent bugs in a single pass (Table~\ref{tab:fixer-bugs}); no subsequent iteration has been lost to the agent-coordination layer.

\begin{table}[tbp]
  \caption{Bugs caught by the Code Fixer in its first deployment (single pass over the Researcher's proposed \texttt{train.py}). The DDP wrapper bug alone would have wasted 10+ hours running an 8-GPU job on a single device.}
  \label{tab:fixer-bugs}
  \centering
  \begin{tabular}{lc}
    \toprule
    Bug category & Count \\
    \midrule
    DDP / multi-GPU misconfiguration       & 3 \\
    GPU memory mismanagement (OOM precursors) & 4 \\
    Unnecessary recomputation / no-op layers & 3 \\
    Distributed dataloader / sampler bugs   & 2 \\
    Eval-loop / metric-aggregation errors   & 2 \\
    \midrule
    Total                                   & 14 \\
    \bottomrule
  \end{tabular}
\end{table}

\begin{figure*}[t]
  \centering
  \includegraphics[width=0.85\textwidth]{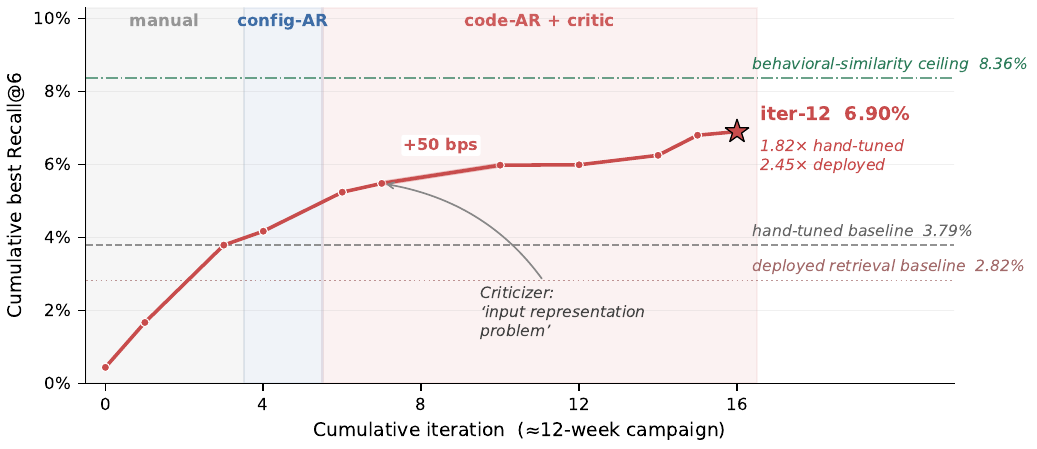}
  \caption{System~A cumulative best Recall@6 across ${\sim}$12 weeks. Background shading distinguishes three regimes: manual engineering, configuration-level AutoResearch, and code-level AutoResearch with the three-agent framework. The Criticizer's ``input representation problem'' directive precedes the largest improvement attributable to a single directive (+50\,bps). The peak (6.90\%, iter-12) is a 1.82$\times$ lift over the hand-tuned baseline.}
  \label{fig:system-a-progression}
\end{figure*}

\subsection{Redirection: Criticizer}

The Criticizer monitors for stagnation and emits strategic directives. It fires only when the best Recall@6 has not improved by more than a relative threshold over a window of recent iterations, so most iterations proceed without intervention; when it does fire, it produces a 2--5 sentence directive whose role is to suggest a different region of the design space rather than a specific code change. The Researcher consumes the directive on the next iteration and translates it into code. The directive is cleared on the next Keep, so the Criticizer never accumulates an authoritative voice across the campaign---it acts only as a stagnation breaker. The relative-gain threshold (8\%) and window size (5 iterations) were chosen empirically: 3 iterations was too frequent (firing on noise); 8 iterations was too late (compute already wasted).

The most consequential directive of the System~A campaign came after five iterations of diminishing returns fusing pre-compressed features:

\begin{quote}
\itshape\raggedright
\textcolor{black!75}{[Criticizer, Run~6, after iteration~4]}\\
The gap is not a modeling problem---it is an input representation problem. Build explicit co-purchase neighborhood representations.
\end{quote}

\noindent The Researcher then built multi-hop neighborhood profiles, driving the trajectory from 5.48\% to 5.98\% Recall@6 in three iterations---the largest improvement we attribute to a single Criticizer directive. Subsequent AutoResearch iterations built on this multi-hop foundation, reaching 6.90\% at iter-12 (Figure~\ref{fig:system-a-progression}). \textbf{Human \texttt{program.md} updates} (System~B) play the analogous role between runs: a human-driven Phase~3$\to$4 pivot redirected exploration from content-only embeddings toward behavioral+content embeddings, ultimately producing System~B's best result (WC\,=\,0.735). The pattern is consistent across both systems: the agent optimizes effectively within a paradigm, but paradigm shifts require external redirection---automated for cheap-to-emit directives, human for cross-run programs.

\subsection{Persistence: Cross-Job Memory}

We extend Karpathy's \texttt{program.md} mechanism in two directions. First, we move it to durable cross-job storage so each new training job inherits the accumulated discoveries of all previous jobs (System~A: 6 runs spanning ${\sim}$10 weeks). Second, in System~B the human researcher (rather than the agent) maintains the program file between runs, encoding constraints, revert thresholds, and known findings the agent should not re-discover. The program's effectiveness is partial: the agent re-tested 100-epoch training at least six times across multiple runs despite the program explicitly prohibiting it. The mechanism reduces the frequency of memory decay but does not eliminate it.

System~B also adds \textbf{automatic revert}, which triggered on 4 of 21 iterations in one run when the agent explored configurations that caused OOM failures. The revert is cheap to use because System~B's iterations cost on the order of dollars; the same approach is impractical for System~A where a single iteration costs hundreds.

\begin{figure*}[t]
  \centering
  \includegraphics[width=0.85\textwidth]{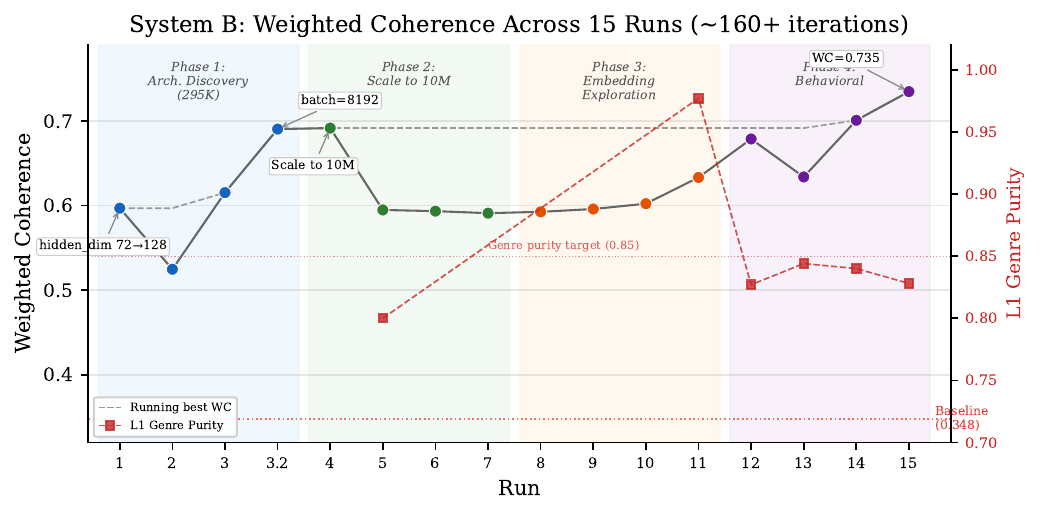}
  \vspace{-2mm}
  \caption{System~B weighted-coherence progression across four phases (15 runs, 150+ iterations). Architectural decisions discovered in early runs lock in for all subsequent runs; the human-driven Phase~3$\to$4 pivot from content-only to behavioral+content embeddings produced the campaign best (WC=0.735).}
  \label{fig:system-b-progression}
\end{figure*}

\subsection{Metric Fixation: Partial Mitigation}

Three practices reduced the severity of metric fixation but did not solve it. \textbf{Multi-metric reporting} expanding beyond the primary metric allowed the agent to observe and incorporate secondary signals, although a single optimization target remained; \textbf{usability context} in \texttt{program.md} provided explanations of why each metric mattered downstream; and \textbf{human QA between runs} diagnosed metric fixation early on and mitigated with competing criteria. These helped the human researcher minimize metric-gaming, but the fundamental problem remained that the agent itself never \textit{questioned} the metric. While this is by design (the agent can only modify training parameters, not the evaluation criteria), it makes it impossible for the system to run completely hands-free. Whether this is solvable within current LLM capabilities is, in our view, an open problem (Section~VI).

\section{Results}

\subsection{System~A: Recall@6 Trajectory}

The hand-tuned baseline was the product of four engineer-weeks of manual work. \textbf{Configuration-level AutoResearch}---the same paradigm but with the agent restricted to hyperparameter and architecture edits---plateaued around baseline, unable to surpass the manually tuned result. \textbf{Code-level AutoResearch with the full three-agent framework} reached a \textbf{1.82$\times$} relative lift over the hand-tuned baseline (3.79\%) and \textbf{2.45$\times$} over the deployed retrieval baseline (2.82\%) on its peak evaluation date ($\pm$\,0.28\,pp across 11 evaluation dates; the peak is ${\sim}$2.6$\sigma$ above the across-date mean), an iter-12 result obtained over six AutoResearch runs spanning ${\sim}$10 weeks, within the broader twelve-week campaign that also includes the manual-engineering phase shown in Figure~\ref{fig:system-a-progression}. Even the across-date mean preserves a 1.63$\times$ lift over the hand-tuned baseline.

The agent's progressive discoveries included cross-attention, sentence-transformer fine-tuning, hard-negative contrastive learning~\cite{chen2020simclr,robinson2021hardneg,vandenoord2018cpc}, a temperature curriculum, multi-hop neighborhood profiles, and a dual mean+top-K aggregation that drove the iter-12 lift. The agent additionally produced a 5.8$\times$ catalog-coverage scale-up via an autonomously designed text-only fallback, which we describe separately in Section~V-B. Total compute: 60+ iterations. We report this as an offline study; an online pilot of the embeddings is underway (Section~VI).

\textbf{Component contribution.} A fully randomized A/B over the multi-week 
campaign was not feasible; the natural campaign transitions in Table~\ref{tab:ablation} serve as the ablation. Three lines of evidence support a 
causal rather than coincidental reading. First, the Code Fixer's contribution is independently documented: on its first deployment it corrected 14 latent bugs in a single pass (Table~\ref{tab:fixer-bugs}), including a DDP wrapper that would have run an 8-GPU job on a single device for 10+ hours---a concrete, iteration-level effect separable from any trend in the metric trajectory. Second, the Criticizer's largest impact is sequentially documented: the ``input representation problem'' directive (Run~6, iteration~4) precedes the largest improvement attributable to a single directive (+50~bps over the next three iterations) with no intervening architectural change; the improvement is a spike, not a drift. Third, the Researcher-only configurations bracket the hand-tuned baseline but cannot escape it: configuration-level AutoResearch plateaued just below baseline (3.73\%, 0.98$\times$), and even single-agent code-level AutoResearch---which first cleared the baseline---stalled at 4.17\% (1.10$\times$) regardless of iteration count. The ceiling was broken only as each additional agent was introduced (Reviewer/Fixer to 1.45$\times$, Criticizer to 1.82$\times$), establishing that the gain came from the specific agents rather than accumulated iterations alone.

\begin{table}[t]
  \caption{Effect of each agent on System~A. Lift is measured against the 3.79\% hand-tuned baseline. Rows correspond to natural campaign transitions.}
  \label{tab:ablation}
  \centering
  \small
  \begin{tabular}{lcc}
    \toprule
    Configuration & Best R@6 & Lift \\
    \midrule
    Hand-tuned baseline (no agents)            & 3.79\%  & 1.00$\times$ \\
    Researcher only (config-level AR)          & 3.73\%  & 0.98$\times$ \\
    Researcher only (code-level AR)            & 4.17\%  & 1.10$\times$ \\
    Researcher + Code Fixer (no Critic)        & 5.48\%$^{*}$ & 1.45$\times$ \\
    Researcher + Fixer + Criticizer (full)     & \textbf{6.90\%} & \textbf{1.82$\times$} \\
    \bottomrule
  \end{tabular}
  \\[2pt]
  {\footnotesize $^{*}$Stagnation point after which the Criticizer fired. Pre-Critic best.}
\end{table}

\subsection{System~A: Autonomous Coverage Scale-Up}

The trajectory in Figure~\ref{fig:system-a-progression} reports Recall@6 on the held-out evaluation set, but it understates a second result: midway through the campaign, the agent independently identified and solved a catalog-coverage gap that the hand-tuned baseline had left open. The behavioral co-purchase graph covers only ${\sim}$17\% of items---only the items with at least one observed co-purchase edge---leaving ${\sim}$83\% of items in the broader catalog without any embedding. The hand-tuned baseline simply omitted these items from the FAISS index, capping retrieval at ${\sim}$17\%.

In an iteration whose primary objective was an unrelated loss-function experiment, the Researcher noticed the coverage gap and proposed a text-only fallback: items absent from the behavioral graph would receive an embedding inferred from their text features alone, projected into the same 128d space as the behavioral embeddings via a small MLP trained jointly with the main model. The Code Fixer's pre-execution review caught one DDP wrapper bug in the fallback path; the iteration ran successfully and the next FAISS index covered all items. Coverage thus expanded by \textbf{5.8$\times$} in a single iteration, without any human direction. No human had specified the gap as a target; no \texttt{program.md} entry mentioned coverage.

This is the cleanest example in our data of AutoResearch operating beyond hyperparameter search. Several practitioner implications follow. First, scope expansion is qualitatively different from metric optimization: the agent did not improve Recall@6 (it adjusted; coverage and recall trade against each other), but it removed a structural limitation of the system. Second, autonomous scope expansion is only useful when the agent has access to the metadata that exposes the gap---in our case, the coverage signal was available because the program included data-pipeline statistics. Without that signal, the agent would have continued optimizing on the ${\sim}$17\% subset indefinitely. Third, the Code Fixer is essential here: the fallback path introduced a non-trivial DDP wrapper change, the kind of latent infrastructure fault that would otherwise have wasted the iteration entirely.

\subsection{System~B: Phased Progression}

System~B exhibits the same redirection pattern as System~A, but with the human---not an automated Criticizer---issuing the cross-run pivot.

\begin{table}[t]
  \caption{System~B phased progression across 15 runs (150+ iterations).
  Phase transitions are human-driven \texttt{program.md} updates; gains
  within each phase are agent-driven.}
  \label{tab:sysb-phases}
  \centering
  \begin{tabular}{llcc}
    \toprule
    Phase & Trigger / Focus & Best WC & Lift \\
    \midrule
    Baseline & Hand-tuned (2--3 days)              & 0.348  & 1.00$\times$ \\
    Phase 1  & Agent: architecture discovery        & 0.691  & 1.98$\times$ \\
    Phase 2  & Human: scale to 10M + genre purity   & 0.692  & 1.99$\times$ \\
    Phase 3  & Human: content-only embedding pivot  & 0.633  & 1.82$\times$ \\
    Phase 4  & Human: behavioral+content pivot      & 0.735  & 2.11$\times$ \\
    \bottomrule
  \end{tabular}
\end{table}

Table~\ref{tab:sysb-phases} and Figure~\ref{fig:system-b-progression} show the per-phase progression. The agent
drove the largest single gain autonomously (Phase~1: 0.348~$\to$~0.691), 
discovering key architectural decisions---bottleneck encoder, deeper 
encoder, batch-size and codebook scaling---that locked in for all 
subsequent phases. Phase~3 regressed (0.692~$\to$~0.633), reflecting 
two compounding factors: content-only embeddings had reached a 
representational ceiling, and the genre purity constraint introduced 
in Phase~2 prevented the agent from recovering weighted coherence 
through codebook scaling alone---the same metric-gaming path that had 
been explicitly guardrailed. This regression established that neither 
embedding type nor metric weighting could resolve the 
coherence--purity tradeoff, freeing the human researcher to pivot 
toward behavioral signal integration in Phase~4, producing the 
campaign best (WC~$=$~0.735). The pattern mirrors System~A: the agent 
optimizes effectively within a paradigm; paradigm shifts require 
external redirection---automated in System~A, human-driven in 
System~B.

Equally valuable was what the agent \textit{ruled out}. The three usability criteria (coherence, genre purity, median cluster size) were never simultaneously met in 150+ iterations, indicating that the coherence-density tradeoff is structural, not a tuning problem. This finding freed the human researcher to design a deterministic post-processing solution outside the model. Across 150+ iterations, ${\sim}$65\% were productive; the remainder was waste from infrastructure failures, re-tests, and marginal tweaks. Due to iterative cost running low for System~B, this was not a major concern.

\subsection{Cross-System Findings}

The two systems differ in per-iteration cost by nearly three orders of magnitude, yet exhibit the same failure modes---all four cost- and horizon-driven modes appear in both, and metric fixation appears wherever evaluation is multi-criteria (System~B). This is the strongest evidence in our data that the failure taxonomy is structural rather than artifactual. The systems also share a common breakthrough pattern: the agent reaches a local optimum by incremental search, then external redirection (automated in System~A, human in System~B) reframes the problem and unlocks the next regime.

\subsection{Validation Protocol}
We took three precautions to confirm the reported gains are real rather than artifacts. First, \textbf{robustness across evaluation dates.} A single lucky evaluation date could flatter the result, so we re-evaluated System~A's iter-12 model on 11 distinct customer-purchase dates; the gain held throughout (peak 6.90\%, mean 6.18\% $\pm$ 0.28\,pp, with the peak ${\sim}$2.6$\sigma$ above the across-date mean). Second, \textbf{an immutable evaluator.} In both systems the agent can read the evaluator's output but cannot change its code, so it cannot inflate its own score by gaming the metric. Third, \textbf{full-scale re-testing} (System~B). To rule out that the gains were an artifact of cheap, small-scale exploration, each architectural decision found in early runs (bottleneck encoder, deeper encoder, batch-size and codebook scaling) was re-tested at the larger Phase~2 data scale before being locked in for later phases.

\section{Discussion}

\textbf{Where AutoResearch delivers value.} Despite the failure modes, AutoResearch delivered concrete value: \textit{throughput} (150+ experiments in three weeks for System~B; autonomous overnight runs for System~A); \textit{systematic coverage} that proved the coherence-density tradeoff was fundamental rather than a tuning problem; \textit{correct abandonment} of dead ends; \textit{transfer of findings} across runs; and \textit{autonomous problem-solving} (System~A's 5.8$\times$ coverage scale-up; Section~V-B). We estimate AutoResearch saved approximately 8 engineer-weeks across both systems. This estimate sums the engineer-equivalent of wasted System~A iterations (18\% of 60+ at 9--19 hours each), memory-decay re-tests across both systems, and the manual exploration cost that 220+ agent-driven experiments would have demanded at standard ML research rates.

\textbf{Iteration cost as a design parameter.} The two systems differ in per-iteration cost by orders of magnitude, and the scaffolding that made AutoResearch viable differed accordingly. The high-cost system benefits from a pre-execution Code Fixer that catches multi-hour bugs before resource allocation. The low-cost system relies on post-execution automatic revert that lets the agent attempt cheap hypotheses freely and rolls back on regression. This is a two-data-point empirical observation rather than a proven design principle, but we offer it as a parameter to consider when extending AutoResearch to other production deployments: \textit{when iterations are expensive, prevent; when iterations are cheap, react}.

\textbf{The human role at production scale.} Human involvement shifts from running experiments to designing evaluations, writing research programs, and making strategic pivots---higher-leverage decisions while the agent handles systematic execution. More fundamentally, AutoResearch frees the human to \textbf{move upstream}: while the agent optimizes model parameters, the human focuses on whether the input data is right, whether the evaluation captures what matters, and whether the solution should be inside the model at all. System~B exemplifies this: the agent's 150+ iterations established the coherence--genre-purity tradeoff was structural, freeing the human to design a post-processing solution the agent could never have reached.

\textbf{Practitioner takeaways.} For teams considering AutoResearch deployment, we offer five concrete recommendations distilled from the campaign:
\begin{enumerate}
\item Identify the smallest self-contained experimental loop (hypothesis $\to$ run $\to$ evaluation) that still produces meaningful signal. AutoResearch amplifies whatever loop it is given; over-broad loops waste budget, over-narrow loops cannot reach interesting improvements.
\item For iterations with non-negligible cost, add a pre-execution semantic gate (Code Fixer). The amortized savings of catching one DDP wrapper bug pay for the gate.
\item Persist findings to durable cross-job storage from day one. \texttt{program.md} is not just a Karpathy artifact---it is the only mechanism that survives context-window resets.
\item Add a stagnation-only redirection signal. A Criticizer that fires every iteration produces noise; one that fires only on plateau produces paradigm shifts.
\item Proactively treat metric fixation as an unsolved problem. Design multi-metric reporting, encode usability context in the program, and budget human QA between runs. Do not expect the agent to question the metric.
\end{enumerate}

\textbf{Limitations.} Three limitations bound the scope of our 
conclusions. First, the component ablation for System~A is 
observational: the campaign transitions in Table~\ref{tab:ablation} 
are natural rather than controlled, and we cannot fully rule out 
that accumulated iterations---rather than specific agent 
interventions---contributed to the trajectory. While we address this through sequential and artifact-level evidence (Section~V-A), a
controlled replication on a different system would strengthen the 
causal claim even further. Second, both systems operate within a single domain (book recommendation) at a single organization; whether the five failure modes and the cost-dependence principle generalize to other domains remains to be established. We discuss how the three principles map to other agentic domains under \textit{Beyond AutoResearch} below, but empirical validation outside recommendation is future work. Third, the evaluation metrics (Recall@6, weighted coherence) are offline proxies that were chosen for the evaluation of these specific systems; neither has been validated against actual customer outcomes in this study, and the gap between offline improvement and production impact is a known risk in recommendation systems. Both metrics were chosen for their established correlation with the online metrics used in this pipeline, and online evaluation through the production A/B testing framework is the standard deployment path in our setting: online pilots of the System~A embeddings and the System~B Semantic IDs are underway, and validating the offline-to-online translation is the immediate next step for this work.

\textbf{Open problems.} Four problems remain open and motivate future work: \textit{cost-aware iteration planning}---the agent currently weighs no information-gain-vs-compute tradeoff; \textit{parallel exploration} across iterations to hedge multi-hour bets; \textit{cross-system agent communication} (System~A's embeddings feed System~B's input, but the agents are disconnected); and \textit{metric self-evaluation}, the most fundamental open problem---teaching agents to question whether the metric is right requires a form of meta-reasoning that current LLMs do not reliably perform.

\textbf{Beyond AutoResearch.} The three principles described here---prevent, persist, and redirect---are not specific to ML research automation. They address structural properties that arise in any long-horizon autonomous agent loop: actions with asymmetric failure cost (motivating a pre-execution gate), context resets between sessions (motivating durable external memory), and search spaces with local optima (motivating stagnation-triggered reframing). The same triad applies to agentic software engineering, where a pre-execution semantic gate catches destructive migrations before CI/CD commits them, a cross-session memory preserves architectural decisions across coding sessions, and a stagnation signal redirects an agent cycling on a failing test rather than reconsidering the root cause. It applies equally to automated scientific experimentation, where preventing a failed wet-lab run is orders of magnitude more valuable than reverting it, accumulated findings must survive personnel turnover, and null-result patterns need to trigger hypothesis-level pivots rather than parameter-level adjustments. The cost-dependence insight transfers directly: the weight placed on each principle should scale with the irreversibility of actions, the length of sessions, and the dimensionality of the search space---not applied uniformly across all deployments.

\section{Conclusion}

The most important lesson from twelve weeks of production-scale AutoResearch is about the agent's boundaries. Both systems exhibit the same collaboration pattern: the agent exhaustively maps within-model limits, and the human designs around them. In System~A, the Criticizer's directive---\textit{``build explicit neighborhood representations''}---drove the largest improvement we attribute to a single directive, a paradigm shift the Researcher could not reach through incremental code search. In System~B, 150+ iterations established that the coherence--genre-purity tradeoff was structural, freeing the human to design a deterministic post-processing solution the agent could never have proposed. The three-agent framework---pre-execution prevention, cross-job persistence, and stagnation-triggered redirection---is what makes this collaboration pattern viable at production scale, where the cost of an unguarded iteration is measured in hours and hundreds of dollars. AutoResearch does not replace the researcher. It replaces the researcher's weekends---and frees them to work on the problems that matter most.

\section*{Reproducibility}

The orchestrator design (Researcher / Code Fixer / Criticizer agents, stagnation thresholds, sliding-window context management) and the evaluator protocol are described in Section~IV and Section~V. Hyperparameter values for the Researcher, Code Fixer, and Criticizer (relative-gain threshold 8\%, stagnation window 5 iterations, model assignments \texttt{claude-sonnet-4-6} for the Researcher and \texttt{claude-opus-4-6} for the Code Fixer and Criticizer) are stated in the body. The training data---co-purchase behavioral graphs and customer purchase logs---is proprietary and cannot be released; we describe the schema (anonymized customer--item interactions, text features from a pretrained sentence transformer~\cite{reimers2019sbert}) and the evaluation protocol (FAISS~\cite{johnson2019faiss} IVF+Flat top-6 retrieval against held-out next-purchase) in sufficient detail that the framework can be replicated on other behavioral datasets. The complete reproducibility checklist required by the ICDM Applied Track is submitted alongside this manuscript.

\section*{Acknowledgment}

\textit{AI-Generated Content Disclosure (IEEE Policy).} The work reported in this paper concerns large language models acting as autonomous research agents; LLMs are therefore the \emph{subject} of the paper. The agents described here---Researcher, Code Fixer, and Criticizer---are powered by Anthropic's Claude (\texttt{claude-sonnet-4-6} for the Researcher; \texttt{claude-opus-4-6} for the Code Fixer and Criticizer) accessed through Amazon Bedrock. The agents autonomously generated, modified, and executed training-script code across 220+ experiments; this is the contribution being studied and is described in the body of the paper (Sections~II--V). Separately, in preparing this manuscript the authors used a large language model (Anthropic Claude) for editorial assistance: tightening prose, organizing structure, and identifying inconsistencies between sections. All technical content, factual claims, experimental results, and arguments are the authors' own and were verified against source experimental records. No LLM was used to generate or analyze experimental data, write the orchestrator code described in this paper, or produce results reported herein.

\bibliographystyle{IEEEtran}
\bibliography{AutoResearch_at_Production_Scale_ICDM2026_CR}

\end{document}